# Viia-hand: a Reach-and-grasp Restoration System Integrating Voice interaction, Computer vision and Auditory feedback for Blind Amputees

**Chunhao Peng[1], Dapeng Yang[1,2,*], Ming Cheng[1], Jinghui Dai[1], Deyu Zhao[1], Li Jiang [1]**

1 State Key Laboratory of Robotics and System, Harbin Institute of Technology, Harbin 150080, China;

2 Artificial Intelligence Laboratory, Harbin Institute of Technology, Harbin 150080, China

* Corresponding author (yangdapeng@hit.edu.cn)

**Abstract:** Visual feedback plays a crucial role in the process of amputation patients completing grasping in the field of prosthesis control. However, for blind and visually impaired (BVI) amputees, the loss of both visual and grasping abilities makes the "easy" reach-and-grasp task a feasible challenge. In this paper, we propose a novel multi-sensory prosthesis system helping BVI amputees with sensing, navigation and grasp operations. It combines modules of voice interaction, environmental perception, grasp guidance, collaborative control, and auditory/tactile feedback. In particular, the voice interaction module receives user instructions and invokes other functional modules according to the instructions. The environmental perception and grasp guidance module obtains environmental information through computer vision, and feedbacks the information to the user through auditory feedback modules (voice prompts and spatial sound sources) and tactile feedback modules (vibration stimulation). The prosthesis collaborative control module obtains the context information of the grasp guidance process and completes the collaborative control of grasp gestures and wrist angles of prosthesis in conjunction with the user's control intention in order to achieve stable grasp of various objects. This paper details a prototyping design (named viia-hand) and presents its preliminary experimental verification on healthy subjects completing specific reach-and-grasp tasks. Our results showed that, with the help of our new design, the subjects were able to achieve a precise reach and reliable grasp of the target objects in a relatively cluttered environment. Additionally, the system is extremely user-friendly, as users can quickly adapt to it with minimal training.

**Keywords:** transradial prosthesis, visual impairment, computer vision, auditory aid, human-machine interaction

## I. Introduction

Human prehension skills are developed through interactive learning, with vision and touch being crucial[1]. Blind and visually impaired (BVI) amputees have lost their visual and partial tactile perception abilities, making it difficult for them to achieve the reach-and-grasp task by manipulating the prosthesis hand.

Existing prosthesis control methods rely heavily on amputee visual perception and require patients to approach and aim at objects using environmental perception and system supervision of human vision, which is extremely difficult for BVI amputees. Prosthesis control methods based entirely on electromyographic (EMG) signals, such as finite state machine[2], [3], pattern recognition[4]–[6], synchronous control[7]–[9], and deep learning[10]–[12], require patients to align the prosthesis with the object before performing corresponding EMG control. The new shared control method based on multi-sensor fusion can achieve a certain degree of automation of prosthesis control through environmental context information [13]–[15], particularly the automatic grasp estimation method based on computer vision[16], [17]. For example, MYO-PACE[18]detects the relative state of the prosthetic hand and the object using global computer vision and realizes continuous decoding of the user's intention, allowing the dexterous prosthesis hand and wrist to be controlled by man-machine cooperation. However, the above-mentioned man-machine cooperative control method is still based on the assumption that patients actively complete environmental perception and object positioning through human vision, which is incompatible with BVI amputees.

Blind guidance, that is, guiding patients to move near the objects (environmental perception and mobile navigation), as well as accurately driving prosthesis to aim at objects (object positioning), is a prerequisite for BVI amputees to complete the grasp task. The key to guiding the blind is to assist BVI in obtaining accurate and timely information from their surroundings in order to improve their perception and understanding of their surroundings. The key to the blind guidance task is to help the BVI obtain information accurately and quickly from the environment to facilitate their perception and understanding of the environment. In addition to retinal prosthesis technology with high risk and high cost[19]–[21], sensory compensation provided by hearing or touch is the main idea of blind guidance research. For example, pedestrian navigation system based on voice interaction[22], [23], auxiliary robot [24], [25] or moving cane[26] based on vibration tactile, and a tongue display unit (TDU) applied to the tongue [27], etc. The above system helps the BVI complete mobile navigation by transmitting the environment information (including text, scene, product, etc.) detected by the camera to the BVI through voice description and tactile feedback (mechanical vibration or electrical stimulation).

Mobile navigation tasks typically direct users to positions on a two-dimensional plane. Finding objects in 3D space is required for object location. The use of wearable computer vision devices to help the BVI locate and grasp the object in the environment through auditory feedback (voice prompts[28], [29], "radar" prompts [30], image-to-sound feature mapping [31]–[33]) has been proven effective. Moreover, the method of continuous proximity feedback has better positioning effect[34]. In the latest research, Hu et al. [35] utilized the ability of humans to perceive the position of sound sources and used spatial audio rendering to improve the efficiency and accuracy of object positioning. Furthermore, some researchers use a cooperative robot system to provide hand guidance[36] or vibration tactile sense [37], [38] to assist BVI in completing close object operations. The hand of BVI patients (tactile sense [39]) plays a key role in the above technology, and the above method will have significant limitations for amputated BVI individuals.

Aiming at BVI amputees who have lost both their visual and grasping abilities, this paper proposes a system framework that combines voice interaction, environmental perception, grasp guidance, prosthesis collaborative control, auditory feedback, and tactile feedback, in order to realize the reach-and-grasp restoration of BVI amputees. Our proposed system has obvious advantages in mobile navigation and grasp guidance over the traditional scheme, such as natural, efficiency, stability, and accuracy. Furthermore, the further improvement direction of the BVI prosthesis system is proposed on the basis of fully obtaining the experimental data of the experimenters and according to the personal feelings of the subjects.

## II. Methods

### A. Framework

Faced with the needs of BVI amputees, as well as prosthesis control and blindness guidance research, this paper proposes the framework of a man-machine collaborative multi-sensory closed-loop BVI prosthesis system, as shown in Figure 1. The framework combines computer vision, auditory/tactile feedback, collaborative prosthesis control, and voice interaction. It primarily consists of six functional modules: 1) voice interaction module 2) environmental perception module 3) grasp guidance module 4) prosthesis collaborative control module 5) auditory feedback module 6) tactile feedback module.

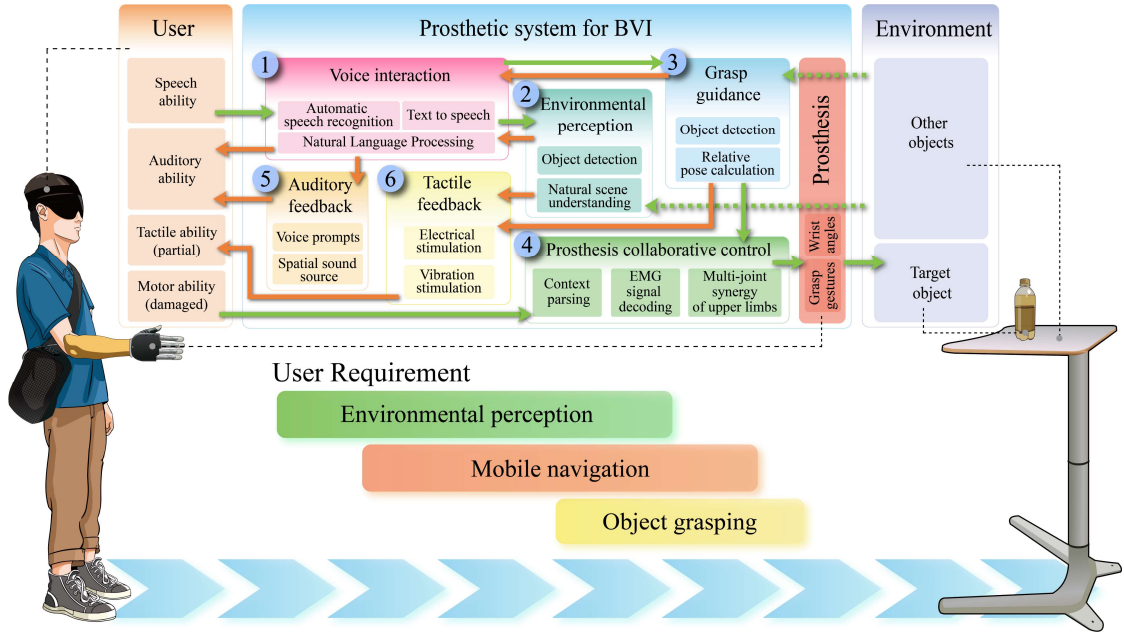

Fig. 1 The framework of BVI prosthesis system

The modules collaborate with each other and with users to form a closed-loop human-computer interaction system capable of realizing functions such as environment perception, mobile navigation, object grasping, and so on, as well as meeting the basic needs of BVI amputees. In this case, the voice interaction module receives user instructions and activates other functional modules accordingly. The environmental perception and grasp guidance module gathers environmental data using computer vision and relays it to users via auditory and tactile feedback modules. The prosthesis collaborative control module also obtains the contextual information (including object category, object posture, and the relative position between the object and the prosthetic hand) and combines the user's control intention (EMG signal decoding and upper limb multi-joint coordination) to complete the collaborative control of the prosthesis grasp gestures and wrist angles, resulting in the stable grasp of different objects.

**B. Prototyping**

Based on the foregoing framework, a prototype of a wearable BVI prosthesis system, depicted in Figure 2, is developed in this paper, which includes all functional modules (except the vibration tactile feedback module) depicted in Figure 1. We call this system "viia-hand", as it integrates the voice interaction, two cameras (global and local), and auditory feedback all together ("i" stands for "eye").

The prototype of the system uses intelligent voice interaction, global computer vision for environmental perception, local computer vision to guide patients in object grasping, the multi-fingered dexterous prosthetic hand to grasp objects, and auditory feedback through spatial sound sources and voice prompts. It can help BVI amputees complete perception, positioning, navigation, and grasp tasks in the indoor environment. The hardware of the viia-hand mainly includes a global RGB-D camera, a local RGB-D camera, a microphone, an earphone, a dexterous prosthesis (HIT-VI [40]), a separate sound source module, the main control system, a mobile power supply and a backpack. The software mainly includes automatic speech recognition, text to speech, object detection, wireless communication, and a prosthesis control thread.

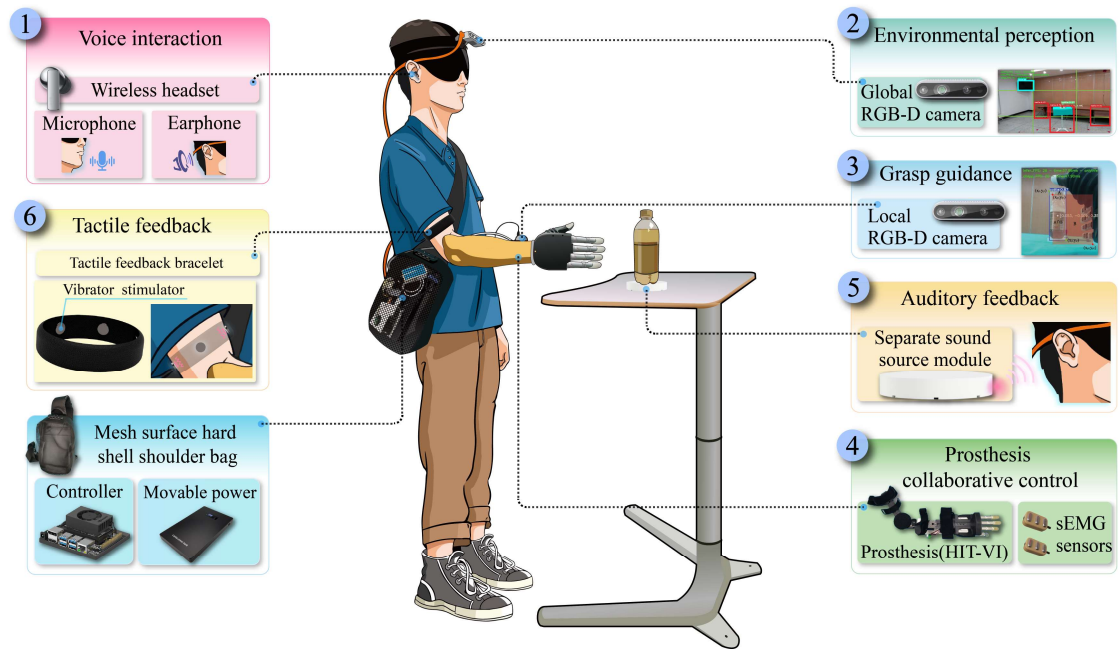

Fig. 2 Schematic diagram of the wearable BVI prosthesis system prototype (viia-hand)

The system includes two RGB-D cameras: 1) The global RGB-D camera worn on the user's head is responsible for achieving environmental perception close to the user's field of vision. 2) The local RGB-D camera installed on the prosthesis's wrist is responsible for accurately locating the position of the target object and achieving prosthesis grasp guidance.

The system provides auditory feedback in two ways: 1) a separate sound source module bound to the target object 2) a voice prompt delivered via earphone. When the separate sound source module is awakened, it emits spatial sound that can assist users in locating objects, and each target object matches a separate sound source module to form a spatial sound source system based on the Internet of Things (IoT).

Furthermore, the user's voice instruction (VI) is input through the microphone device, which is designed in conjunction with the earphone (wireless headset) and worn on the user's ear.

The system adopts NVIDIA® Jetson Xavier™ NX (external dimension 70x45mm, power consumption 10-20w) as the controller, and builds Python multithreading software architecture under Ubuntu platform, which integrates global/local environment recognition and object detection, voice interaction, wireless communication and prosthesis collaborative control.

Given the system's wearability, a mesh and hard shell shoulder bag is designed to provide enough space and heat dissipation for the internal controller and battery (19V 120000mah). We designed a special prosthesis wearing bracket for the experimental object (healthy subjects) in this paper, which is worn on the entire right arm of the user, so that the weight of the prosthesis is shared on the upper arm and the lower arm, reducing the user's use burden.

## III. Experiments

The grasp evaluation experiment of the viia-hand will be introduced in this section. We designed an experiment of object positioning and grasping in a real-world indoor environment and verified the actual effect of BVI amputees grasping objects with the help of the viia-hand.

### A. Subjects

Six healthy subjects (with an average age of about 24 years, one of whom is left-handed) were recruited to participate in the experiment. To make the experimental conditions as close to the actual situation of the end users (BVI amputees) as possible, the subjects did not know the specific composition of the viia-hand and wore an eye mask before entering the experimental site to avoid remembering the indoor layout. Our experiment was approved by the university's ethics committee and followed the Helsinki Declaration.

### B. Experiment set-up

The experimental site of the viia-hand built in this paper is shown in Figure 3. The experimental area is a fan-shaped area with a radius of 5m and an angle of 120 degrees. The center of the fan-shaped area is the starting point of each experiment of the subjects. We divide the edge of the fan-shaped area into 10 segments, which are used to randomly place the experimental table, and the separate sound source module (in this paper, coaster shape) and the grasp target (in this paper, a bottle) are placed on the table.

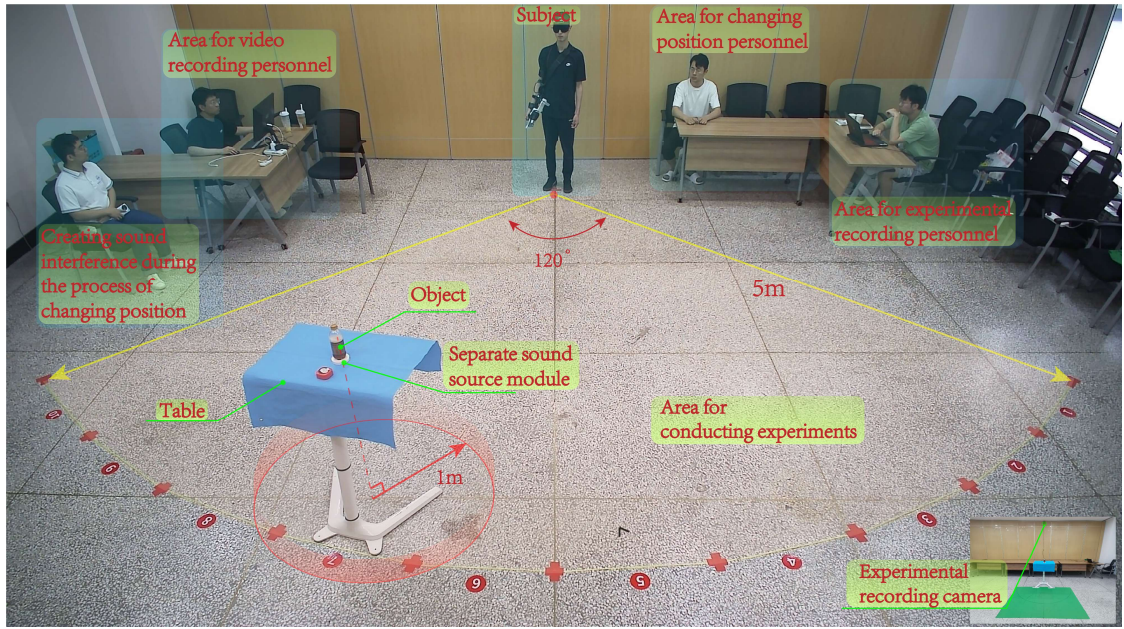

Fig. 3 Composition and layout of experimental site for indoor grasp experiment

## C. Experiment Procedure

The staff blindfolded the subjects outside the experimental site, assisted them in wearing the viia-hand, and transported them to the experiment's starting point indoors. Before the formal experiment starts, it is necessary to introduce the use process of the system and the key steps of the experiment to the subjects, and then start the formal experiment after the subjects understand the use method of the system.

First, the staff chose a location at random from 10 placement points (at least two placement points away from the previous experimental location) and placed the experimental table, bottle, and separate sound source module there.

Following the grasp, the subject controls the opening of the prosthesis, and the system determines whether the grasp was successful based on the sensors on the prosthesis, prompting the subject with "This grasp task is over, grasp is successful" or "This grasp task is over, grasp is failed" based on the grasp results. The staff then assisted the subjects in returning to their starting positions, and the single experiment concluded.

The staff selects the next experimental position and moves the table there. In order to avoid the influence of the sound generated by the table movement on the subject's judgment of the location of the target object, another staff was arranged to make noise interference during the process of moving the table.

Repeat the preceding steps, ensuring that each subject performs 40 grasp experiments and that the subjects get enough rest during the experiment. In each experiment, the experimental recorder records the relevant experimental data, including the reach time $t_1$ (from the system prompt "real-time detection in progress" to " reached the accessible range." in seconds), the alignment time $t_2$ (from the system prompt " reached the accessible range " to " reached the graspable range " in seconds), and whether the grasp is successful (yes or no). At the same time, the system will automatically records the video data of global and local RGB-D cameras during the experiment. The video recording personnel will also save the video data of the camera equipment used for experimental recording in the experimental site.

## IV. Results and Discussions

### A. Analysis of Experimental Results

In this paper, the viia-hand is evaluated from the time to complete the grasp and the success rate of the grasp. The time required for the reach process ($t_1$) and the time required for the alignment process ($t_2$) are counted as indicators of grasp efficiency to verify the adaptation of the subjects to the viia-hand. To count the success rate of the subjects' grasp and verify the actual effect of the viia-hand. In order to facilitate comparative analysis, the experimental results of six subjects are counted in a graph, as shown in Figure 4.

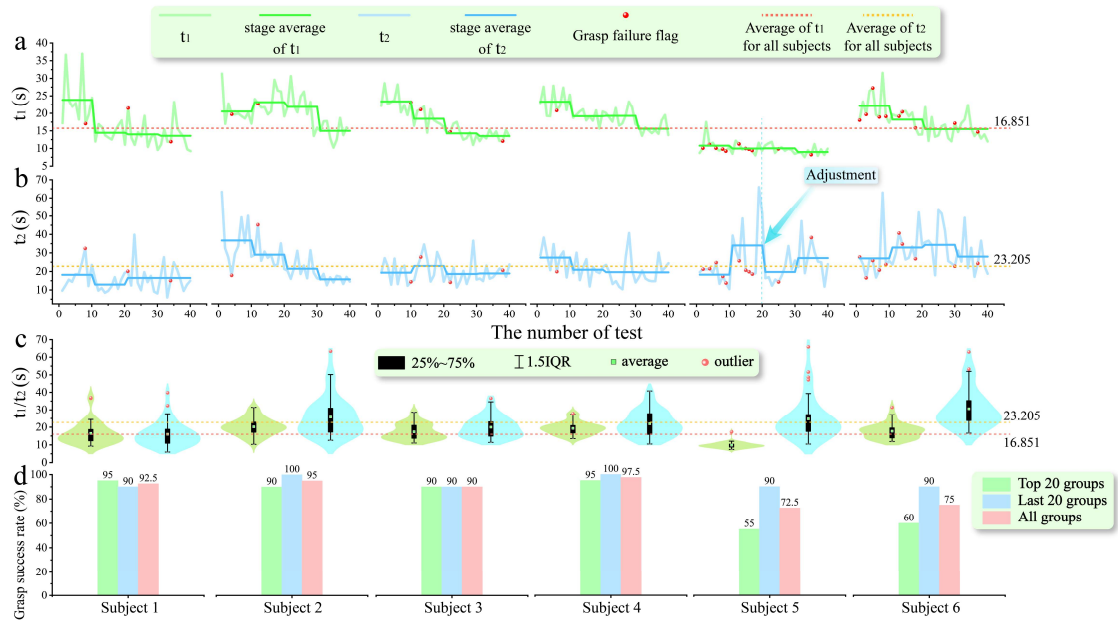

Fig. 4 Experimental data of six subjects a) changes of reach time $t_1$ ) changes of alignment time $t_2$  c) data distribution of reach time $t_1$(green) and alignment time $t_2$(blue) d) success rate of grasp.

We conducted a comparative experiment on two of the six subjects (subjects 5 and 6). The contrast of subject 5 is reflected in two aspects among them: 1) to represent the performance of BVI in a familiar environment, we described the basic layout of the experimental site to subject 5 (other subjects are unaware of the experimental site); 2) subject 5 was only required to hold his arm during the last 20 experimental groups (other subjects were required to hold his arm in each group of experiments. During the experiment's preparation, we discovered that the stability of the arm has a significant influence on grasp, so we instructed the participants to hold the arm wearing the prosthesis with the other hand during the alignment process to ensure the stability of the movement of the prosthesis, in order to verify the specific influence of the movement stability of the prosthesis on the experimental results.). The contrast of subject 6 is reflected in the arm operation habit, whose dominant hand is left hand.

To summarize, the experimental results show that it is difficult for BVI amputees to complete the grasp operation, and the process is influenced by a variety of factors (for example, the familiarity of the site, the stability and accuracy of the action, the personal operating habits and the proficiency in the use of equipment, etc.). The viia-hand proposed in this paper can assist users in completing the tasks of mobile navigation and object grasping in practical application, and users can quickly adapt to and become familiar with the system's working process after short-term training. Through communication with the subjects, they generally believe that the system's navigation, positioning, and alignment methods are natural and efficient, and that after a few simple experiments, the system can be quickly mastered, allowing the grasp task to be easily completed. Furthermore, the subjects praised the system for its wearability and stability.

## V. Conclusion

For BVI amputees, aiming at the tasks of environmental perception and object grasping in living settings, this paper proposes a system framework that combines voice interaction, global environment perception, local grasp guidance, human-robot shared control, and auditory/tactile feedback. Based on this, we have developed a wearable prosthesis system for BVI(viia-hand). To verify the design, we set up an indoor experimental scene, and carried out a grasp experiment of the viia-hand. The results show that the viia-hand can help BVI amputees complete the tasks of environmental perception, mobile navigation, object positioning and grasping in indoor scenes, and the subjects all agree with the wearing comfort, running stability and natural efficiency of the system. Through the exploration of this article, further improvement directions for the BVI prosthesis system are proposed to efficiently complete grasp operations for BVI amputees. The aim is to provide a safe and reliable wearable device for BVI amputees to help them complete some activities of daily life (ADLs).